\documentclass{svproc}

\usepackage{microtype}
\usepackage{subfigure}
\usepackage{booktabs} 
\usepackage{mathtools,amssymb}
\usepackage[ps,dvips,matrix,arrow,frame,import,curve,color]{xy}
\usepackage{graphicx, hyperref, caption,xcolor}

\usepackage{amsthm}
\usepackage{centernot}
\usepackage{url}

\newtheorem{thm}{Theorem}[section]
\newtheorem{prop}[thm]{Proposition}
\def\eq#1{Eq. \ref{eq:#1}}

\newcommand{\be}{\begin{equation}}
\newcommand{\ee}{\end{equation}}
\newcommand{\ba}{\begin{array}}
\newcommand{\ea}{\end{array}}
\newcommand{\bea}{\begin{eqnarray}}
\newcommand{\eea}{\end{eqnarray}}
\theoremstyle{remark}
\newtheorem{rmk}[thm]{Remark}
\newcommand{\R}{\mathbb {R}}

\newcommand{\Sphere}{\mathbb{S}}
\newcommand{\CH}{\mathcal {H}}
\newcommand{\CR}{\mathcal {R}}
\newcommand{\CT}{\mathcal {T}}
\newcommand{\CL}{\mathcal {L}}

\newcommand{\hitsone}{{\tt hits@1}}

\newcommand{\norm}[1]{{\lVert {#1} \rVert}}
\newcommand{\dimRE}{{\dim_{RE}}}

\newcommand{\TransE}{{\sc TransE}}
\newcommand{\PairRE}{{\sc PairRE}}
\newcommand{\RotatE}{{\sc RotatE}}
\newcommand{\ComplEx}{{\sc ComplEx}}
\newcommand{\DistMult}{{\sc DistMult}}

\newcommand{\RE}{{\sc RE}}
\newcommand{\wikikg}{{\sc ogbl-wikikg2}}

\definecolor{avocado}{RGB}{86, 130, 3}
\newcommand{\omri}[1]{}

\newcommand{\michael}[1]{\textcolor{purple}{{Michael S: #1}}}

\newif\iflongversion
\longversionfalse

\begin{document}
\mainmatter              

\title{What is Learned in Knowledge Graph Embeddings?}
\titlerunning{What is Learned in Knowledge Graph Embeddings?}

\author{Michael R. Douglas\inst{1,2,6} \and 
Michael Simkin\inst{1} \and
Omri Ben-Eliezer\inst{1,3} \and
Tianqi Wu\inst{1,4} \and
Peter Chin\inst{1,5} \and
Trung V. Dang\inst{5} \and
Andrew Wood\inst{5}
}
\authorrunning{Michael R. Douglas et al.} 

\institute{
CMSA, Harvard University, Cambridge MA, USA,\\
\email{mdouglas,msimkin@cmsa.fas.harvard.edu},\\
\and
Dept. of Physics, YITP and SCGP, Stony Brook University, Stony Brook NY, USA\\
\and
Department of Mathematics, MIT, Cambridge MA, USA, \email{omrib@mit.edu}\\
\and
Depts. of Mathematics and Computer Science, Clark University, 950 Main Street Worcester MA, USA,
\email{tianwu@clarku.edu}
\\
\and
Department of Computer Science, Boston University, Boston MA, USA,
\and
IAIFI, MIT, Cambridge MA, USA
}

\maketitle

\begin{abstract}
A knowledge graph (KG) is a data structure which represents entities and relations as the vertices and edges of a directed graph with edge types. 
KGs are an important primitive in modern machine learning and artificial intelligence.  Embedding-based models, such as the seminal \TransE\  
[Bordes et al., 2013] and the recent {\PairRE} 
[Chao et al., 2020] are among the most popular and successful approaches for representing KGs and inferring missing edges (link completion).   Their relative success is often credited in the literature to their ability to learn logical rules between the relations.

    In this work, we investigate whether learning rules between relations is indeed what drives the performance of embedding-based methods.  We define motif learning and two alternative mechanisms, network learning (based only on the connectivity of the KG, ignoring the relation types), and unstructured statistical learning (ignoring the connectivity of the graph). 
    Using experiments on synthetic KGs, we show that KG models can learn motifs and how this ability is degraded by non-motif (noise) edges.
    We propose tests to distinguish the contributions
    of the three mechanisms to performance, and apply them to popular KG benchmarks.
    We also discuss an issue with the standard performance testing protocol and suggest an improvement.  \footnote{
To appear in the proceedings of Complex Networks 2021.
    }

\end{abstract}

\section{Introduction}

\subsection{ Definitions and basic properties of KGs }
\label{ss:defkg}

A knowledge graph (henceforth abbreviated KG) is a graph-structured
data model, often used to store descriptions of entities such as
people, places, and events.
A KG can be defined as a collection of ordered triples 
$(h,r,t) \in V\times T\times V$, where $V$ is the set of entities 
and $T$ is the set of relation types.   
As an example, consider the relation ``John lives in Chicago''; here $h = \text{\rm ``John''}$,  $r=\text{\rm ``lives\, in''}$, and $t=\text{\rm ``Chicago''}$.

In graph theoretic terms, a KG is
a directed graph in which each edge has both an orientation and a type
(or label).  Its vertices correspond to entities, and each triple $(h,r,t)$
corresponds to an edge.  We will also use the notations $h \xrightarrow[]{r} t$,
$t \xleftarrow[]{r} h$ and $(t,-r,h)$ to denote the triple $(h,r,t)$,
and $h\centernot{\xrightarrow[]{r}} t$ to denote absence of a triple.

\iflongversion
\else
KGs are more flexible than traditional database models, but more structured
than text, which facilitates automated reasoning.  
Many represent  ``general knowledge,'' including Wikidata, 
YAGO \cite{pellissier2020yago} and DBpedia \cite{lehmann2015dbpedia}.
There are many more KGs covering specialized topics, such as
SemMedDB for biomedical data \cite{kilicoglu2012semmeddb}.
\fi
KGs have also been created as benchmarks to test KG software.  The KGs
in our experiments came from the Open Graph Benchmark (OGB)
collection \cite{hu_open_2020} and the OpenKE compilation \cite{han2018openke}.
The largest of these is the {\wikikg} benchmark.
It is based on Wikidata, and has 17,137,181 edges of 535 types
connecting 2,500,604 entities.

\subsection{ Embedding models }
\label{ss:embed}

Much of the success in learning KGs is due to \textit{embedding models}. Such models identify the vertices with points in a metric space with and interpret proximity under a relation-type-dependent transformation as graph adjacency (with the corresponding edge label). 
The literature describes many embedding models for knowledge graphs, including: 
\TransE\ \cite{bordes_translating_2013}, \RotatE\ \cite{sun_rotate_2019}, 
\PairRE\ \cite{chao_pairre_2020}. For simplicity and concreteness in this work we primarily consider {\TransE} and {\PairRE}, due to their state-of-the-art performance. 
While the models are similar that similar reasoning can be applied to their analysis, they differ in expressiveness, as highlighted in \cite{chao_pairre_2020}. For results on the expressive power of {\TransE}, we refer the reader to \cite{bhattacharjee2020relations}.

In what follows, $G$ denotes a KG with vertex set $V$, edge types $T$, and relations $R$. Both {\TransE} and {\PairRE} consist of an underlying vertex map $\psi : V \to \Sphere^{k-1}$.\footnote{ Many implementations of {\TransE} embed into $\R^k$.
Also, the norms below vary but are often $\ell_1$.}
They differ in their approach to modelling relations.
 
In {\TransE} for each $r \in T$ the model learns a vector $\mu_r \in \R^k$. The idea is that for every triple $(h,r,t)$, it holds that $\psi(h)+\mu(r) \approx \psi(t)$ if and only if $(h,r,t)$ is a relation in $G$.

In {\PairRE} for each $r \in T$ the model learns two vectors $\mu_r^h,\mu_r^t \in \R^k$. The idea is that $\mu_h(r) \circ \psi(h) \approx \mu_t(r) \circ \psi(t)$ if and only if $(h,r,t)$ is a relation in $G$. Here $\circ$ denotes the Hadamard  (elementwise) product $(a\circ b)_i = a_i b_i$.

KG embedding models are trained using standard ML techniques, by gradient descent
on the error with which these defining properties hold.  One slightly nonstandard
point is that negative sampling is used to estimate the error for
the triples not in $G$ (negative edges or simply negatives).
Recent works use self-adversarial sampling \cite{sun_rotate_2019} in which the negatives are sampled by making use of the model being trained. 

As detailed in the next section, there are two common applications for KGs: Binary link classification and ranking of potential link completions. For the classification task, the models learn a threshold $\gamma > 0$. A triple $(h,r,t)$ is predicted to be in $G$ if and only if $\norm{\psi(h) + \mu_r - \psi(t)} < \gamma$ (in the case of {\TransE}) or $\norm{\mu_r^h \circ \psi(h) - \mu_r^t \circ \psi(t)} < \gamma$ (in the case of {\PairRE}). In the ranking task the model is given a head $h \in V$, a relation $r \in T$, and a list of potential completions $t_1,\ldots,t_n \in V$. The model ranks the likelihood of the relations $\{(h,r,t_i)\}_{i=1,\ldots,n}$ in increasing order of $\norm{\psi(h)+\mu_r-\psi(t_i)}$ or $\norm{\mu_r^h\circ\psi(h) - \mu_r^t\circ\psi(t)}$ for {\TransE} and {\PairRE}, respectively.

\subsection{ Tasks and evaluation }
\label{ss:tasks}

The most studied KG task is link completion:
given partial information about a relation, say its
type and the head, find the ``valid'' tails (or given the tail and type, find the heads).
We stress that in contrast to link prediction in network theory,
the type of the relation is important.

\iflongversion
\else
What does ``valid'' mean?  If we take it to mean edges which are already in the 
graph, then this is a problem of database retrieval.
\fi
One could instead consider link completion as a particular type of knowledge graph completion.  This refers to tasks such as filling in
missing entries in a KG, correcting errors, predicting the time evolution of a KG, or other forms of inference, all with the common assumption that there is an implicit ``ground truth'' KG to which the dataset is an approximation.  A simple example would be a dataset which is a sample from a known KG.  In this case, the
valid completions would be those which appear in the complete KG.
The hypothesis is that the KG has structure which can be learned and used to make
these predictions. 

Here are some examples of structures in Wikidata which could be used to make link
predictions:\footnote{
Lists and descriptions of the Wikidata entities and relations are readily
available, 
try for example {\tt https://www.wikidata.org/wiki/Q5}}
\begin{itemize}
    \item Property {\tt P47}, ``shares border with,'' relates pairs of geographic regions.
    It is symmetric, so from $A\xrightarrow{P47} B$, we can deduce $B\xrightarrow{P47} A$.
    \item Property {\tt P103}, ``native language,'' implies property {\tt P1412},
    ``language(s) that a person speaks, writes or signs, including the native language(s).''
    \item Property {\tt P131}, ``located in the administrative territorial entity,''
    is transitive.
\end{itemize}
We will refer to structures of this type, which impose relations between the relations,
as ``rules,'' and discuss them more systematically below.

As a more complicated example, by combining {\tt P937}, ``work location,'' which relates people and
places, with {\tt P37}, ``official language,'' which relates places and languages,
we could hope to deduce
{\tt P1412}, ``languages spoken, written or signed.''  This would be a statistical
rather than a logical inference, but a very likely one.  A further complication is that
{\tt P937} links are supposed (by the Wikidata guidelines) to be as specific as possible,
so we might need to use {\tt P131} to make the inference as well.  One can see that there
is a large scope for this type of inference, and that the number of rules,
each of which would need to be programmed in a traditional approach, is also large.  The
prospect of automatically learning these rules is very attractive.

To evaluate these ideas, one needs a test dataset of triples and a measure of link
completion accuracy.  One can of course split a larger dataset into training and
testing sets by sampling.  While this is very standard in ML, for a prediction
problem it can be criticized on the grounds that the model can take advantage
of structure only visible once the completions are known.  To avoid this criticism,
the {\tt ogbl-wikikg} benchmark did its split by sampling
the Wikidata information at three different dates, and then using the links
added during the two intervals as the validation and test datasets.
This has the potential problem that the data addition process might be nonstationary (time dependent).  

The most popular measure of accuracy is defined as follows (for tail prediction; head prediction is analogous).  
For each testing triple $(h,r,t)$, we give
$h$ and $r$ to the model, which gives us a list of candidates for $t$ ranked by score.
The {\tt hits@N} metric is then the fraction of triples for which the correct $t$ has rank
$N$ or higher, and the {\tt MRR} (mean reciprocal rank) metric is the mean of $1/\mbox{rank}$
over the test set.

For large KGs the full list of candidates for $t$ is expensive to evaluate, so in practice one often considers a subset.
This is usually chosen by filtered uniform sampling, meaning that candidate completions $t'$ are uniformly sampled from the vertex set
excluding those for which the testing or training dataset contains the triple $(h,r,t')$.  The OGB
benchmark evaluates both head and tail completion, each with 500 filtered negatives, and reports the results for the combined test set.

Looking at the OGB leaderboard\footnote{
{\tt ogb.stanford.edu/docs/leader\_linkprop}} 
and Table \ref{tab:results},
the {\wikikg} 
dataset can be completed using {\TransE} (with 500 dimensional embeddings)
to get testing {MRR} 0.43 and {\tt Hits@1} 0.41.  In other words, without
knowing anything about the relations other than the graph,
given an entity and relation type, this simple model can predict in
over 40\% of cases the other entity involved in the relation.  Taken at face
value, this is remarkable.
How does this work?  What structure in the dataset is being used?

\subsection{ Do KG models learn rules? }
\label{ss:mech}

A popular hypothesis is that the KG models are learning 
rules which are usually satisfied by the relations.  The
simplest rules involve pairs of links:
symmetry ($A\xrightarrow[]{1} B \Leftrightarrow B\xrightarrow[]{1} A$), exclusivity ($A\xrightarrow[]{1} B \Rightarrow 
A \centernot{\xrightarrow[]{2}} B$), 
and subrelation ($A\xrightarrow[]{1} B\Rightarrow A\xrightarrow[]{2} B$).  
Other rules involve
multiple links, such as the conjunction rule,
\be\label{eq:conjrule}
A\xrightarrow[]{1} B\xrightarrow[]{2} C \Rightarrow
 A\xrightarrow[]{3} C .
\ee

Many KG works advocate a model by showing its ability to express these rules.
For example, in {\TransE} the rule \eq{conjrule} is naturally expressed by the
embedding property
\be\label{eq:transe-conj}
\mu_1 + \mu_2 \approx \mu_3 ,
\ee
as then $\psi(B)\approx \psi(A)+\mu_1$ and $\psi(C)\approx \psi(B)+\mu_2$ 
will imply $\psi(C)\approx \psi(A)+\mu_3$.

Now it is not {\it a priori} obvious that KG completion is operating by learning and
using these rules.  There might be other structures in the dataset, such as the
clustering which is much studied in graph theory, which are responsible.  It might
also be that while rules can be learned in principle, the real world KGs do not have
high enough signal to noise to do this.

Let us preview some experiments which bear on these questions 
(see \S \ref{s:experiments} for details):
\begin{itemize}
    \item Use {\TransE} for the KG completion task, but learn only the vertex embeddings and freeze the relation embeddings.  This gets almost the same MRR as the original model.
    
    \item Replace all the edge labels (in both training and testing datasets)
    with a single label.   Now the MRR drops, but only from 0.43 to 0.36.

\end{itemize}
Since we expect these modifications to drastically handicap rule learning,
such results cast doubt on the idea that rule learning by learning properties
such as \eq{transe-conj} is the main explanation of KG model performance. 

\subsection{ An issue with evaluating large KGs }
\label{ss:issue}

Before we take these unexpected results too seriously, we should ask to what
extent they might be explained by problems with the data
or evaluation procedures.

The practice of using a sampled list of negatives is a shortcut which does not correspond to a real KG task,
so it should be justified by comparison with the ``true metrics'' computed using a complete list.
As we will see in \S \ref{s:experiments}, while 500 negatives is adequate for our other KGs,
it is quite small for {\wikikg}.  We noticed this by evaluating a baseline 
(or ``null'') model which 
(for tail completion) ignores the head and takes the score of $(h,r,t)$ to be the conditional probability $P(t|r)$ estimated on
the training data as
\be \label{eq:pgdef}
P(t|r) \equiv P(\mbox{tail}|\mbox{rel}) \sim \frac{\mbox{Number of edges }x\xrightarrow[r]{} t \,\forall x}{\mbox{Number of edges }x\xrightarrow[r]{} y \,\forall x,y}.
\ee
This simple model gets an MRR of $0.23$ for tail prediction.  

To understand why, consider 
the relation {\tt P1412},
``languages spoken, written or signed.''
There are about 250 entities in Wikidata
which represent languages, of which only a few are common.
But since there are about $2.5\cdot 10^6$ vertices, the probability that a uniform sample of size 500 will contain even a single
language entity is $250\cdot 500/2.5\cdot 10^6 \sim 0.05$.  So, with very high probability, the correct result will rank first,
just because it is the only language entity on the list.\footnote{
It also turns out that {\tt P1412} is over-represented in the test
set.  In all, it contributes about $.10$ of the total
tail MRR=0.23 of the simple model.}
Looking at all the items in
the {\wikikg} test set, only about 10\% of the entries have any negative vertices with the correct relation type. 
This suggests that the true metrics could be rather different.
We could still use these uniformly sampled metrics to compare models, if they are monotonic in the true metrics.
Even so, one might lose discriminatory power. 

These general observations are not new to us; in \S \ref{ss:related} we cite 
several works which point out the need for the testing procedure to use plausible negatives and
propose ways to get them, using either human input or another
inference procedure to create the negatives.
A direct but costly way to solve the problem is to increase $N_{neg}$,
in this example by a factor somewhat larger than 20.

A simpler way to mitigate the problem, new so far as we know, is to sample the negatives using the model \eq{pgdef}.   
In \S \ref{s:experiments} we use a 50-50 mixture of this sampling
with uniformly sampled negatives, and compare these resampled results
with uniform sampling.
As an example, the {\wikikg} resampled (or R-) MRRs
for {\TransE} and {\PairRE} are
0.13 and 0.28 respectively.

On re-evaluating the results from \S \ref{ss:mech}, we find that freezing 
relation embeddings reduces the R-MRR from 0.13 to 0.09, 
and removing edge labels reduces it to  0.06.  So this is part but not all
of the resolution.

\subsection{ Summary of our contributions }
\label{ss:contrib}

Our main contribution is to propose a way to study the question 
``What do KG models learn?''.
We define three types of learning and propose tests to
distinguish their contributions to performance.  Motif learning is a precise definition of the rule learning posited
in many KG works, which depends only on structures in the KG.
Network learning is based purely on connectivity, and unstructured statistical learning uses a graphical model which ignores network structure.

We find this distinction useful for several reasons.  
First, it clarifies the interpretation of experiments.  The standard benchmark KGs have different statistical properties
and this is reflected in different potential performance for the three learning mechanisms.  Rather than say that one KG is
better than another (after all the goal is to work with general KGs), we can factor out these differences and make a combined
interpretation of results.  As for the unexpected results,
their interpretation is clearer once one realizes that the link completion
task can be solved in different ways.
Second, network learning and graphical models are classic topics and are far better understood than the general
problem of KG learning.  By seeing how they fit into the general problem, we make a principled start on bringing the general
theory up to the same level.  For example, we can use the mathematics of graph embeddings to understand network learning.

By study of synthetic KGs, we show that popular KG embedding models can do motif learning, and study how this degrades with
noise and other features of the problem.  
The mathematics of graph theory predicts a phase transition at a critical noise threshold,
and we exhibit this.
We can also study freezing relation learning
in a controlled setting and argue that the remaining performance is due to network
learning.

\subsection{ Related work }
\label{ss:related}

Given the relative success of embedding-based methods for knowledge representation, there have been many works explaining the efficacy of these methods from various perspectives; here we describe some representative and closely related works. 
\cite{pezeshkpour2020revisiting}  and \cite{sun-etal-2020-evaluation} point out that current metrics for evaluating KG methods have significant flaws and suggest alternative evaluation procedures. 
\cite{allen_understanding_2019} aim to understand the latent structure of knowledge graph embeddings by leveraging insights from word embeddings. They import semantic concepts from the natural language processing literature, such as paraphrases, analogies, and context shifts, and find evidence that these concepts also play a role in some relation types of KG embeddings. 
\cite{kadlec-etal-2017-knowledge} and~\cite{jain2020knowledge} demonstrate that simple baselines can sometimes outperform much more complicated KG embedding techniques, which is in line with many results in this paper. \cite{Akrami2020} suggest that many of the most popular benchmarks used to evaluate embedding methods contain significant redundancies and are thus not sufficiently challenging to capture the difficulties arising with real-world data. \cite{chandrahas-etal-2018-towards} initiate an investigation of the geometry of different types of embedding methods.

\omri{While I wrote in detail only about the most relevant work, I should make an effort to make the list of references a bit longer by also referring to works that are slightly less relevant (but not irrelevant) -- at this point the list is shorter than what people usually do in these conferences.}

\section{ Three mechanisms of KG learning }
\label{s:questions}

Besides learning rules, what other structure could the KG models be using?
Let us state two alternate hypotheses, and then restate rule learning as
motif learning, a definition which only uses KG structure.  For all three,
their precise definition will be in terms of a restriction on the information
which can be used in the mechanism.

\subsection{ Unstructured statistical learning }
\label{ss:stats}

A simple first hypothesis is that the models are not learning the network structure, rather they are picking up on statistical information such as \eq{pgdef}.
There are many more sophisticated models of this type, 
such as \cite{sutskever2009modelling}.
A broad class are covered by
\begin{definition}
Unstructured statistical learning models the probability distribution
of triples $(h,r,t)$ in terms of latent variables $l_h,l_r,l_t$ as
\be
P(h,r,t) = \sum_{\substack{l_h\in\CL_h\\ l_r\in\CL_r \\ l_t\in\CL_t}}
P(h|l_h) P(r|l_r) P(t|l_t) P(l_h,l_r,l_t).
\ee
\end{definition}
This is a standard graphical model \cite{bishop2006pattern} which can
easily learn constraints of this type, but cannot learn rules or network structure.%
\footnote{
The perceptive reader will note that as stated this is false, with the simplest
counterexample being to identify $\CL_h\cong\CH$ and take 
$P(h|l_h)=I[h=l_h]$, {\it etc.}.
It is surprisingly difficult to make this constraint precise, and we
plan to do this elsewhere.
For present purposes we approximate it by requiring $|\CL_h|\ll|\CH|$, {\it etc.}.
}
There are variants which can learn symmetry, subrelation and
exclusivity in terms of joint probabilities of relations with the
same head and tail.  

How far can this idea go towards explaining link prediction results ?
We will discuss the general model of this type elsewhere.  
If we take the latent
variables to be class probabilities, these look rather similar to embedding
models such as {\sc TuckER} \cite{balazevic_tucker_2019}.

Here we consider an embedding version of the ``null model''
\eq{pgdef}, which we call RE.\footnote{
Following the convention in which KG embedding models have names ending with
the letters capital R and/or E.}
It has separate embeddings
$\psi_h$ for heads, $\psi_t$ for tails and $\mu_r$ for relations.  The score
of a tail completion $(h,r,x)$ is simply $\psi_t(x)\cdot\mu_r$ (or a
normalized version of this).  
Good performance of this model may tell us more about a dataset than about KG
learning, but this illustrates the idea.

\subsection{ Network learning }
\label{ss:prox}

Our next hypothesis is that the models are using the network structure, but only its connectivity, ignoring the relation types.\footnote{
We use the term ``network'' rather than ``graph'' at this point to reduce confusion
with statistical terminology, in particular ``graphical models.''
}  
This certainly seems to fit with the results in \S \ref{ss:mech}!  

For example, if there are many candidates for a tail vertex, 
a network model might prefer the ones closest to the head.
Arguably the simplest definition of ``closest'' is the vertices which minimize
the number of edges in the shortest path, independent of orientations.  
One could propose other definitions, assigning lengths to edges which might depend on node degrees and/or orientation.  This type of proximity structure is easily captured by an embedding model, indeed the topic of metric and similarity embeddings of graphs is very well developed, with many reviews including \cite{heinonen2012lectures}.
We certainly expect that the KG embedding models use proximity as a factor, but to what extent does proximity explain their performance?

As a straw man hypothesis, suppose that we make the predictions by uniformly sampling the distance two neighborhood.  This does very poorly for two reasons.  First, the degree two neighborhood of a KG regarded as an undirected graph tends to be very large, because of the presence of tail nodes of very high degree.\footnote{
While the average degree of a node in {\tt ogbl-wikikg} is 12,
the highest degree node connects to almost 9\% of the other nodes.
It is {\tt Q5}, ``human,'' due to relations such as ``Albert Einstein is a human.'' 
}
This might be dealt with by redefining proximity to exclude such nodes, but even a neighborhood of size $(\mbox{average degree})^2$ is too large for this to work by itself.
However, the combination of proximity with unstructured statistical learning might not be a bad model.  Can we define its separate contribution?

To make this precise, we make the following definition.
\begin{definition}
    Network learning can use any directed graph structure
    which does not depend on edge types.
\end{definition}
This includes degree distributions, distance distributions, spectral properties
and even frequencies of motifs defined without regard to
edge types.  A variant would further restrict
to undirected graph structure.

We can then compare models allowed to use both network and unstructured information,
with those using either separately.

\subsection{ Inference of rules by learning motifs }
\label{ss:motifs}

From a graph theoretic point of view many rules (though not all, for example
disjunction) are related to motifs, small labeled digraphs which appear as subgraphs of the KG.  For example, the symmetry rule is related
to the motif consisting of both orientations of an edge.
The conjunction rule
\eq{conjrule} is related to a triangle motif, a graph with the three vertices
$A,B,C$ and the three directed edges corresponding to the three relations.
Denote the triangle graph with these edge types as
\be\label{eq:conjmotif}
T(1,2,-3) \equiv \{
A\xrightarrow[]{1} B,\, 
B\xrightarrow[]{2} C,\,
 C\xleftarrow[]{3} A 
 \}.
\ee
Note that this motif does not carry exactly the information of the rule.
Whereas \eq{conjrule} treats $A\xrightarrow[]{3} C$ specially,
one could distinguish one of the other edges to get similar but different rules.
However, if we grant that the various rules related to the motif have similar
probabilities, then the problem of learning motifs will be a good approximation
to that of learning rules.  There are many works on identifying and learning motifs
statistically, with a much studied example being the planted clique problem 
\cite{alon1998finding}.

To make these ideas precise, we make
\begin{definition}
 A $k$-motif model can base its predictions on 
 the statistics of labeled directed subgraphs of the KG with up to $k$
 vertices and on the corresponding neighborhood of the given vertex.
\end{definition}
As an example, a 3-motif model which learns \eq{conjmotif} could identify 
a copy of the LHS of \eq{conjrule} with $h=A$ and $r=1$, and predict the RHS.

\subsection{ Distinguishing the three types of learning }

We just outlined three types of learning -- of unstructured statistics, of network structure, and of motif structure.
There might be other learning mechanisms as well, and architectures
suited to them.  A clear case is disjunction, which is not
a motif.\footnote{Some cases of disjunction can be represented
by sets of motifs.}  KG models which work with disjunction often introduce
other structures such as ``boxes'' in embedding space \cite{ren_query2box_2020,abboud_boxe_2020}.

The mechanisms are not exclusive, 
indeed one could argue that KG embedding models provide elegant
combinations of all three.  
Still, to properly interpret results and judge models, it is useful to distinguish
between them.  For example, attributing performance differences to 
rule learning may be misleading
if the other mechanisms have comparable or larger effects.

We defined the learning types in terms of conditions on the information they can use.
Now {\TransE}, {\PairRE} and the other embedding models do not satisfy any of the three conditions, so all three mechanisms might be important.
Thus, we now ask: how can we distinguish the contributions of these
different mechanisms to the performance of a model?

First, the definitions suggest ablation tests:
\begin{itemize}
    \item By removing labels, we only allow network learning.
    \item By freezing the relation embeddings, we disable most of the proposed mechanisms for motif learning.  
    \item By adding random ``noise'' edges with all new relation types,
    unstructured and motif learning should be hardly affected, while network learning
    should be degraded.
    \item Suppose we remove every aspect of the model which relates heads and tails, say by using separate embedding functions for heads and tails.  This should degrade
    network and motif learning much more than unstructured learning. 
\end{itemize}
In all of these cases, the resulting degradation could be interpreted as a measure of the contribution of the affected mechanisms.
One has to be careful as ablations such as freezing weights
could cause more general degradation, 
say if the initialization values are inappropriate.

Another class of tests is to look at expected properties of the 
embeddings.  Claims such as ``In {\TransE}, the relation \eq{conjrule} is learned by
finding relation vectors which satisfy $\mu_1+\mu_2\sim\mu_3$,'' can be checked directly.  As another example,
in \cite{bhattacharjee2020relations} it is shown how to construct {\TransE} embeddings
(without relation types) by starting with a metric embedding of the
undirected graph obtained by forgetting orientations, and adjoining a dimension.
To the extent that this picture is realized, it clearly shows that {\TransE} is learning the network structure.

\section{ Synthetic KGs }
\label{s:genkg}

One approach to studying the mechanisms involved in KG learning is to analyze the performance of various models on \textit{synthetic} KGs. We present several case studies in which we analyze both performance and specific parameters of the embeddings obtained by analyzing synthetic KGs.

Each synthetic KG we consider has the form $G \cup R$, where $G$ is a (highly structured) deterministic KG and $R$ is a random KG which we think of as noise. In all our examples we take $R = \cup_{r \in T} R_r$, where $T$ is the set of relation types and each $R_r$ is an independent random directed graph on $V(G)$ with each edge present independently with probability $p_r \in [0,1]$, and all edges have relation type $r$.

For each graph $G \cup R$ we study the link completion task as discussed earlier with $N_c=500$ for both head and tail prediction. For the training set we take $R$ as well as a random $0.8$-fraction of $G$. For the testing set we take the remaining $0.2$-fraction of $G$. We exclude $R$ from the test set because for the distributions we use, test edges of $R$ would be information-theoretically unlearnable.

We did a suite of link completion runs with various models, scanning the embedding dimension and the hyperparameter $\gamma$. Results quoted are the best train and test MRR and minimal embedding dimension required for this test MRR result.

\subsection{ Link completion without motif learning }


\begin{table}[htb!]
    \centering
    {%
    \begin{tabular}{|c|c|c|c|c|c|c|c|c|}
    \hline
    graph & $G_1$ & $G_1$ & $G_2$ & $G_2$ & $G_3$ & $G_3$ & $G_4$ & $G_4$ \\
    model & TransE & TransE* & TransE & TransE* & TransE & TransE* & TransE & TransE* \\
    MRR & 0.62 & 0.58 & 0.59 & 0.50 & 0.48 & 0.25 & 0.16 & 0.05 \\
    motif norm & 0.022 
    & 1.45 & 0.063 
    & 1.52 & 0.727 & 1.78 & 0.947 & 1.65\\
    \hline
    \end{tabular}
    {\caption{MRR and motif norm for graphs $G_1,G_2,G_3,G_4$ with models TransE and TransE*. The motif norm of $T(i,j,-k)$ is the ratio of $|\mu_i+\mu_j-\mu_k|$ to the average norm of a translation vector. All norms are $\ell_2$. $G_4$ contains two motifs, and both norms are reported.}\label{tab:syn}}
    }
\end{table}

In our first example we consider graphs with $n = 10000$ vertices, defined as follows: Let $G_1$ be a graph with $2499$ disjoint copies of the triangle motif $T(1,2,-3)$ and the remaining $2503$ vertices isolated. For $i=1,2,3$ and $p \in [0,1]$, let $R_i(p)$ denote the random graph where each (directed) edge is present independently with probability $p$, and all have label $i$. Define:
\begin{align*}
&G_2 = G_1 \cup R_1(1/(4n)) \cup R_2(1/(4n)) \cup R_3(1/(4n)),\\
&G_3 = G_1 \cup R_1(3/(4n)) \cup R_2(3/(4n)) \cup R_3(3/(4n)).
\end{align*}

We also define the graph $G_4$ by letting $G$ be the graph with $999$ and $499$ respective copies of the triangle motifs $T(1,2,-3)$ and $T(4,5,-6)$ (all disjoint), and setting
\[
G_4 = G \bigcup_{i=1,2,3,7,9} R_i(8/(5n)) \bigcup_{i=4,5,6,8,10} R_i(4/(5n)).
\]

To test the extent motif-learning plays a role in modelling these graphs, we trained two embedding models: {\TransE} and a variant, {\TransE}*, in which the relation vectors are frozen at their random intialization (so that the optimization is only over the vertex embedding). We report the results in Table \ref{tab:syn}, which we now discuss.

We see that for {\TransE}, the model succeeds in the edge-prediction task quite well (although performance clearly degrades with noise). We remark that for these examples, an MRR of $\approx 0.64$ is the best one could hope for. This is because for each test edge we only expect the algorithm to make a correct prediction if the other two edges in its motif are included in the training set. The latter event has probability $0.8^2 = 0.64$. Another striking feature is that MRR for $G_4$ is approximately one-quarter of the theoretical maximum, despite it having approximately $20$ times more noise edges than motif edges. Finally, for each of these graphs, the motifs themselves are learned quite well. This is evidenced by the small norm (relative to the average norm of translation vectors) of $\mu_i+\mu_j-\mu_k$ for each motif $T(i,j,-k)$.

We cannot expect {\TransE}* to learn the motifs themselves. Nevertheless, the MRR for the link-prediction task remains quite high. This is perhaps due to alternative learning mechanisms, such as network learning.

\begin{table*}[hbt!]
    \centering
    {%
    \begin{tabular}{crrrrrrrr}
      \hline
dataset & nentity & nrels & nedges & ntri(train) & ntri(all) & RE U/R-MRR & $\Delta m$ & $\Delta R$\\
      \hline
wn11 & 38193 & 10 & 112581 & 10033 & 13420 & 12.7/2.8 & 0.30 & 6.2/4.2 \\
wn18 & 40942 & 17 & 141442 & 25144 & 27111 & 10.4/2.7 & 0.14 &  7.4/16.4 \\
fb15k & 14950 & 1344 & 483142 & 4865600 & 6100776 & 25.3/14.2 & 25.57 & 22.8/51.4 \\
fb15k237 & 14504 & 236 & 272115 & 1867991 & 1898714 & 29.8/17.9 & 1.13 & 18.9/25.0\\
ogbl\_wikikg2 & $2.5\cdot10^6$ & 534 & $1.6 \cdot 10^7$ & 10471397 & 11481634 & 42.3/13.3 & 0.63 & 7.2/6.3\\
       \hline
    \end{tabular}
    }
        {\caption{Statistics of knowledge graphs.
        nedges denotes the training set, and ntri denotes the number of triangle
motifs \eq{conjmotif}, for the training set and for training+test.
The {\RE} column is the MRR for that model, and
the $\Delta$ columns are explained in the text.
           \label{tab:stats}}}
\end{table*}

\begin{table*}[hbt!]
\centering
\begin{tabular}{l|l|llllll}
  \hline
  &  & base & 0.2 & 0.5 & 1.0 & 2.0 & XR \\ 
  \hline
 & ComplEx & 70.3/82.1 & 67.7/60.1 & 61.0/47.0 & 54.0/38.0 & 41.9/27.1 & 27.0/12.0 \\ 
 & DistMult
 & 71.0/75.9 & 63.0/48.4 & 51.0/33.0 & 51.0/33.0 & 41.6/24.5 & 53.0/51.0 \\ 
 fb15k & PairRE & 68.0/81.0 & 68.3/78.0 & 68.0/72.0 & 67.0/62.0 & 63.7/52.5 &  50.4/40.1 \\ 
 & RotateE & 68.2/83.0 & 68.2/75.7 & 66.0/64.0 & 66.0/64.0 & 56.5/52.8 & 50.0/38.0 \\ 
 & TransE & 68.0/81.0 & 69.2/80.7 & 70.0/78.0 & 70.0/74.0 & 70.0/74.2 & 46.0/30.0 \\
   \hline
 & ComplEx & 51.0/42.0 & 48.1/33.0 & 44.0/28.0 & 36.0/22.0 & 23.3/14.2 & 31.0/12.0 \\ 
 & DistMult & 51.0/41.0 & 47.0/29.5 & 42.0/25.0 & 38.0/21.0 & 29.1/16.0 & 26.0/13.0 \\ 
 fb15k237 & PairRE & 49.0/38.0 & 47.9/35.7 & 47.0/35.0 & 47.0/34.0 & 46.1/29.7 & 30.0/12.0 \\ 
 & RotateE & 50.0/39.0 & 43.8/32.2 & 40.0/28.0 & 33.0/23.0 & 21.3/16.3 & 32.0/13.0 \\ 
 & TransE & 51.0/39.0 & 50.3/38.3 & 51.0/38.0 & 50.0/38.0 & 50.4/37.7 & 32.0/14.0 \\ 
  \hline
 & ComplEx & 28.0/28.0 & 20.8/20.7 & 15.0/15.0 &  9.0/ 8.0 &  5.6/ 3.7 & 20.0/20.0 \\ 
 & DistMult & 27.0/27.0 & 18.1/17.5 & 11.0/11.0 &  7.0/ 5.0 &  3.7/ 2.6 & 20.0/20.0 \\ 
 wn11 & PairRE & 21.0/15.0 & 17.4/11.8 & 17.0/10.0 & 20.0/13.0 & 15.2/ 6.7 & 25.0/19.0 \\ 
 & RotateE & 28.0/20.0 & 18.3/11.6 & 12.0/ 8.0 &  8.0/ 5.0 &  4.1/ 3.1 & 21.0/17.0 \\ 
 & TransE & 33.0/22.0 & 24.5/11.9 & 21.0/ 8.0 & 17.0/ 6.0 & 14.5/ 4.3 & 27.0/18.0 \\ 
   \hline
 & ComplEx & 92.0/96.0 & 90.7/95.8 & 90.0/95.0 & 89.0/93.0 & 81.2/81.1 & 89.0/93.0 \\ 
 & DistMult & 91.0/96.0 & 90.3/94.6 & 89.0/93.0 & 86.0/85.0 & 60.9/54.5 & 90.0/93.0 \\ 
 wn18 & PairRE & 81.0/80.0 & 89.2/89.2 & 90.0/94.0 & 90.0/94.0 & 79.7/69.4 & 89.0/88.0 \\ 
 & RotateE & 72.0/66.0 & 90.1/93.9 & 90.0/94.0 & 89.0/93.0 & 88.4/90.8 & 87.0/85.0 \\ 
 & TransE & 88.0/88.0 & 89.5/91.6 & 89.0/91.0 & 87.0/87.0 & 76.9/64.1 & 80.0/72.0 \\ 
   \hline
\end{tabular}
\caption{Uniform/resampled MRR by dataset and model (in percent).
The columns are the baseline, varying numbers of noise edges (expressed as a fraction
of the number of training set edges), and XR is the graph without relation types.}
\label{tab:results}
\end{table*}

\subsection{Motif factor with noise}

\begin{figure}
    \centering
    \includegraphics[scale=1]{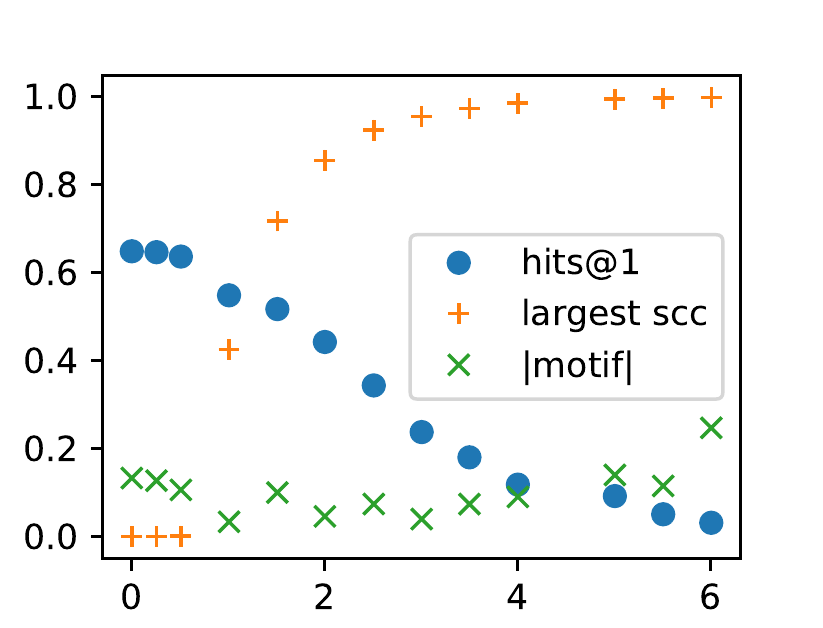}
    \caption{\hitsone{}, largest strongly connected component, and ratio of $|\mu_1+\mu_2-\mu_3|$ to average translation vector norm (in $\ell_1$), as functions of $c$.}
    \label{fig:synth results}
\end{figure}

For our second example we explore the addition of noise to a graph with a strongly represented motif by studying the performance of {\TransE} on $G_1 \cup R_4(c/n)$, for values of $c$ in the interval $[0,6]$. We depict our results in Figure \ref{fig:synth results}.

We first note that for all the graphs, $|\mu_1+\mu_2-\mu_3|$ is small. In other words, the motif $T(1,2,-3)$ is learned. Thus, incorrect predictions are due to the vertex embedding. For $c \leq 0.5$, \hitsone{} is near its maximal value of $0.64$. Above this value, this measure begins degrading. Curiously this phase transition seems to occur together with the appearance of a linear-sized strongly connected component in the underlying undirected graph (see Figure \ref{fig:synth results}). We wonder if this component is difficult to embed, leading to deteriorating performance. Alternatively, perhaps there are \textit{local} explanations for incorrect predictions: As $c$ increases, there are more non-motif triangles (i.e., those including edge label $4$). If proximity and local structure play a substantial role in making predictions, non-motif triangles could cause incorrect predictions.

\begin{table}[htb!]
\centering
\begin{tabular}{lll}
  \hline
   & base & XR \\ 
  \hline
 ComplEx & 48/17.0 & 37/11.6 \\ 
 DistMult & 44/11.0 & 45/14.3 \\ 
 PairRE & 56/28.0 & 39/10.2 \\ 
 RotateE & 43/18.0 & 40/ 9.8 \\ 
 TransE & 44/13.0 & 37/ 6.4 \\ 
   \hline
\end{tabular}
\caption{MRRs for {\wikikg} by model \label{tab:wikikg}}
\end{table}

\section{Experiments on real KGs}
\label{s:experiments}

We used 5 datasets in our experiments, {\wikikg} and 4 KGs from OpenKE \cite{han2018openke}.  Five modified versions of the KGs were considered,
one without relation types (XR), and four with noise.  Much as in \S \ref{s:genkg},
the noise consists of randomly added edges whose number is a specified
multiple from 0.2 to 2.0 of the number of training edges.
Table \ref{tab:stats} contains their general statistics.

In tables \ref{tab:results} and \ref{tab:wikikg}
we give performance results for 5 models
({\TransE}, {\PairRE}, 
{\DistMult} \cite{yang_embedding_2015},
{\ComplEx} \cite{trouillon_complex_2016} and {\RotatE} \cite{sun_rotate_2019}), evaluated using the OGB code.
We did some optimization of the hyperparameter gamma, while  
the hidden dimension was taken large (400 and 1000 for the smaller KGs
and 500 for {\wikikg}).
Standard and resampled MRRs are given together as U-MRR/R-MRR.
Comparing the two, the similarity of their
rank-orderings can be measured by the Spearman rank correlation coefficient.
This is around $0.9$ for the smaller KGs but drops to around $0.6$ for
\wikikg. 

\subsection{ Interpretation }
\label{ss:exp-int}

Do these results fit with the hypothesized three mechanisms?
Let us point out some suggestive patterns.

First, consider the propensity of each KG to each type of learning. 
    KGs with more relations favor unstructured learning,
and this is apparent in the RE model performance.
KGs with more motifs favor motif learning.  In table \ref{tab:stats},
the column $\Delta m$ is the number of triangle motifs which contain test
set edges over the number of test set edges.  
while $\Delta R$ is the difference between TransE base and XR R-MRR.
Presumably, dependence on network learning will show up in
dependence on noise.  This dependence is strongest for {\sc wn11},
which has the fewest relations.

Next, looking at the dependence on the model, {\ComplEx} and {\DistMult}
work best for the KGs with few relations ({\sc wn11} and {\sc wn18}), and
are significantly more affected by noise than the others.  This is
consistent with the idea that they rely more on network learning.

Conversely, the {\PairRE} model is worse than the others for KGs with
few relations.  It is superior only for {\wikikg}, indeed in the resampled
metric it is the only model to convincingly beat {\RE}.  It is also 
 affected by noise (for {\wikikg} at 0.5, MRR=45.8/14.3),
 suggesting that all three mechanisms are in play.

\section{Conclusions}

We propose that KG learning is due to a combination of three mechanisms, namely, unstructured, network and motif learning, and presented results for synthetic and real KGs which illustrate the idea. 

\bibliography{gntm_graphs}
\bibliographystyle{plain}

\end{document}